\documentclass[conference]{IEEEtran}
\IEEEoverridecommandlockouts
\usepackage{cite}
\usepackage{amsmath,amssymb,amsfonts}
\usepackage{algorithmic}
\usepackage{booktabs}
\usepackage{graphicx}
\usepackage{textcomp}
\usepackage{xcolor}
\usepackage{cite}
\usepackage{hyperref}
\usepackage{stfloats}
\def\BibTeX{{\rm B\kern-.05em{\sc i\kern-.025em b}\kern-.08em
    T\kern-.1667em\lower.7ex\hbox{E}\kern-.125emX}}
\begin{document}

\title{AI-Based Multi-View Echocardiographic Image Synthesis using Dynamic Shape Control\\
\thanks{Disclosures: J.F. is supported by Portuguese national funds through the Foundation for Science and Technology (FCT) under grant 2024.03683.BD (DOI: \href{https://doi.org/10.54499/2024.03683.BD}{10.54499/2024.03683.BD}); A.B. is supported by a research grant of the University of Medicine and Pharmacy “Carol Davila”, Bucharest, Romania (Project No. 11906/21.05.2025); A.Y. is supported by a scholarship from the Egyptian Ministry of Research and Higher Education (2021 recipient), by the Belgian Fund for Cardiac Surgery (489708), and by the Frans Van de Werf scholarship 2025. The authors further acknowledge FCT's computational support through the Deucalion HPC infrastructure (2026.03986.CPCA.A1). \par\medskip
\noindent
\copyright~2026 IEEE. Personal use of this material is permitted. Permission from IEEE must be obtained for all other uses, in any current or future media, including reprinting/republishing this material for advertising or promotional purposes, creating new collective works, for resale or redistribution to servers or lists, or reuse of any copyrighted component of this work in other works.}} 

\author{
\IEEEauthorblockN{
João Freitas\IEEEauthorrefmark{1}\IEEEauthorrefmark{2}\IEEEauthorrefmark{3},
Anca Balinisteanu\IEEEauthorrefmark{2}\IEEEauthorrefmark{4},
Ahmed Youssef\IEEEauthorrefmark{2}\IEEEauthorrefmark{5},
Vitor Hugo Pereira\IEEEauthorrefmark{1}, 
Jan D'hooge\IEEEauthorrefmark{2},
Sandro Queirós\IEEEauthorrefmark{3}} \\

\IEEEauthorblockA{
\IEEEauthorrefmark{1}Life and Health Sciences Research Institute, School of Medicine, University of Minho, Braga, Portugal\\
\IEEEauthorrefmark{2}Department of Cardiovascular Sciences, KU Leuven, Leuven, Belgium\\
\IEEEauthorrefmark{3}2Ai - Applied Artificial Intelligence Laboratory, School of Technology, Polytechnic University of \\Cávado and Ave (UPCA), Barcelos, Portugal\\
\IEEEauthorrefmark{4}Department of Cardiology and Cardiovascular Surgery, University of Medicine \\ and Pharmacy "Carol Davila", Bucharest, Romania\\
\IEEEauthorrefmark{5}Department of Cardiovascular Medicine, Suez Canal University, Ismailia, Egypt
}

\IEEEauthorblockA{
Emails: id11569@alunos.uminho.pt, joao.aaraovideirosfreitas@student.kuleuven.be}}

\maketitle

\begin{abstract}
Although automated methods have improved echocardiographic workflows, progress remains constrained by the need for large, well-annotated datasets. As the development of clinical databases is often limited by ethical and logistical challenges, synthetic data generation has emerged as a promising alternative. However, existing approaches commonly rely on left ventricle shape-based supervision or are tailored to specific cardiac views, limiting the generation of additional standard and unconventional imaging planes and the preservation of multi-view consistency across cardiac chambers. To address these limitations, we propose a novel framework for shape-controlled synthesis of multi-view echocardiographic sequences. First, multi-view cross-sections are extracted from dynamic whole-heart meshes. Ultrasound-like appearance is then introduced through a two-stage AI-based pipeline. In Stage 1, an unpaired GAN–diffusion model translates these cross-sections into realistic echocardiographic frames. As this translation is performed independently at the frame level, Stage 2 employs masked video reconstruction to improve temporal coherence and cardiac motion. Evaluation demonstrated high frame-level realism for both stages, while the integration of Stage 2 substantially improved video-level motion quality and temporal consistency. Overall, the proposed framework enables the generation of realistic multi-view echocardiographic sequences from diverse cardiac models, providing a scalable source of annotated data for dataset augmentation and the development of multi-view cardiac assessment methods.

\end{abstract}


\section{Introduction}
Echocardiography is the primary imaging modality for cardiac assessment in clinical practice due to its versatility, high temporal resolution, and broad accessibility. However, its reliability is limited by operator dependence and image quality, leading to variability in measurements and interpretation \cite{b2}.

Automated analysis methods have shown increasing potential to improve the reproducibility and efficiency of echocardiographic workflows by reducing observer dependence and promoting standardized interpretation \cite{b4}. However, their development remains strongly dependent on diverse and well-annotated datasets. In clinical settings, collecting such data is challenging, particularly due to ethical and privacy constraints. Moreover, available datasets are often task-specific, providing annotations for selected views, structures, or measurements.

\begin{figure*}[!b]
    \centering
\includegraphics[width=0.95\textwidth]{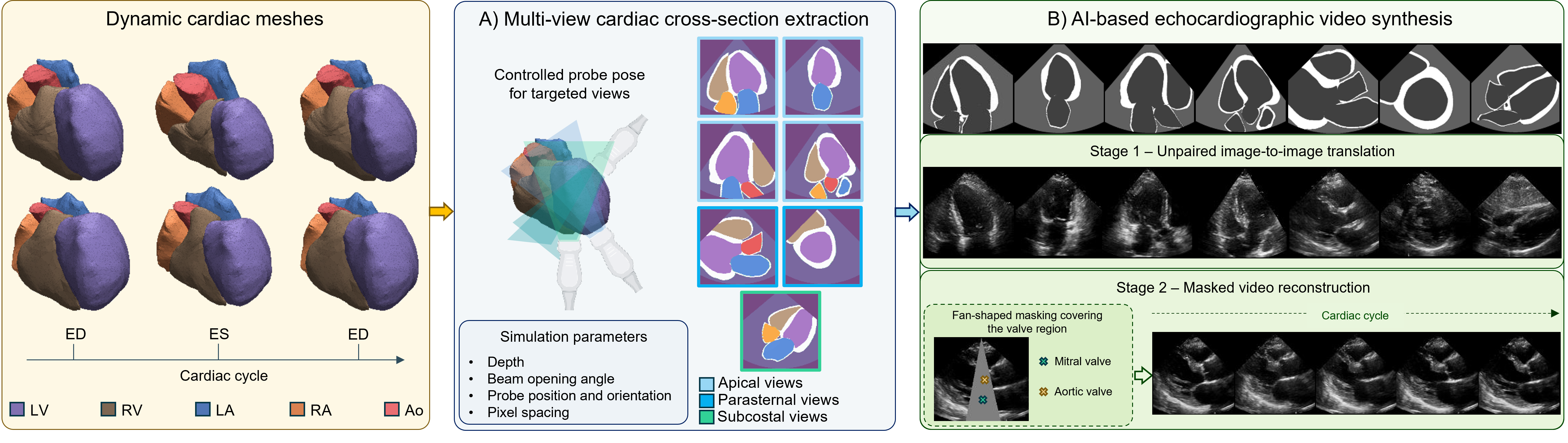}
    \caption{Overview of the proposed framework for synthetic multi-view echocardiographic sequence generation.}
    \label{fig:framework_overview}
\end{figure*}

In this context, synthetic data generation represents an increasingly relevant strategy. Several approaches have been proposed to generate realistic cardiac ultrasound images and sequences with varying anatomy and dynamics. Hybrid simulation and image-translation pipelines enable multi-view synthesis with controlled probe pose \cite{b6}, but their reliance on static whole-heart models limits temporal consistency. Other simulation-based methods generate dynamic sequences \cite{b7}, yet remain anatomically constrained by left ventricle shape-based supervision. More recently, diffusion-based approaches have improved image and video realism \cite{b10} by conditioning on anatomical labels, clinical parameters, or learned motion representations. However, these methods are generally restricted to a limited set of standard views and provide limited control over whole-heart anatomy and imaging plane. Consequently, current approaches do not readily support anatomically controllable whole-heart supervision across multiple views, including other standard or unconventional imaging planes, limiting cross-view consistency and their applicability to comprehensive echocardiographic assessment.

To address these gaps, we propose a framework for shape-controlled synthesis of multi-view echocardiographic sequences from dynamic whole-heart meshes. 


\section{Methodology}

The proposed framework consists of two main modules: (A) multi-view cardiac cross-section extraction; and (B) AI-based echocardiographic video synthesis (Fig.~\ref{fig:framework_overview}). 

\subsection{Multi-view cardiac cross-section extraction}
The first module extracts anatomically consistent cross-sections from dynamic whole-heart cardiac meshes to simulate the acquisition of different echocardiographic views. For this purpose, we employed the ultrasound simulator available in the PLUS Toolkit \cite{b12}, adapting and extending the methodology proposed in \cite{b6} to generate multiclass segmentation-like sequences representing seven echocardiographic views: apical two- (A2C), three- (A3C), four- (A4C), and five-chamber (A5C); parasternal long-axis (PLAX) and short-axis at the papillary muscle level (PSAX-PP); and subcostal (SX).

To mimic the variability commonly encountered during echocardiographic examinations, the probe pose was randomly perturbed around the predefined positions of the perfect views, with offsets sampled from normal distributions following the parameter ranges described in \cite{b6}. In addition, since the apical three- and five-chamber views were newly introduced in this work and are more sensitive to pose perturbations, smaller variations were adopted. Specifically, the standard deviations for rotation, tilt/rock, and translation were set to $(2.5^\circ, 1.5^\circ, 2~\mathrm{mm})$ and $(1^\circ, 1^\circ, 1~\mathrm{mm})$, respectively. Furthermore, the acquisition depth range was extended for the apical views to support focused acquisitions.

\subsection{AI-based echocardiographic video synthesis}
The second module employs a two-stage AI-based framework to transform the anatomical sequences into realistic echocardiographic videos, combining unpaired image-to-image translation with masked video reconstruction.

\vspace{0.2cm}

\textit{Stage 1 - Unpaired image-to-image translation}

\vspace{0.2cm}

In Stage 1, an unpaired image-to-image translation method \cite{b13} was employed to improve frame-wise realism by learning an unpaired mapping between anatomical cross-sections (source domain) and real echocardiographic frames from the MIMIC-IV-ECHO dataset \cite{b14} (target domain), combining adversarial learning with diffusion-based synthesis.

For the source domain, a label-to-intensity mapping was applied to provide clear separation between anatomical regions, assigning distinct intensity values to the background, blood cavities, ultrasound frustum, and cardiac walls.

For the target domain, since MIMIC-IV-ECHO lacks view annotations, the pretrained EchoPrime view-classification head \cite{b15} was used to identify the seven target views, retaining only predictions with a classification probability above 95\%, except for A5C, for which the threshold was reduced to 85\% due to the limited number of available samples. Parasternal predictions were manually curated to exclude focused or very deep long-axis acquisitions and short-axis views at unintended anatomical levels, while subcostal sequences containing inferior vena cava or Doppler acquisitions were removed. Pixels outside the ultrasound field of view were then masked, and a 10\% geometric scaling was applied.

A separate translation model was trained for each view to account for view-specific anatomy and ultrasound appearance. Due to the limited number of A5C sequences, the A5C target domain was supplemented with a subset of A4C samples to reduce overfitting, leveraging their strong visual similarity. During training, source and target frames were randomly sampled from sequences spanning complete cardiac cycles. Although the adversarial and diffusion components were jointly optimized during training, only the adversarial branch output was retained at inference because it better preserved the granular texture characteristic of echocardiographic speckle.

\vspace{0.2cm}

\textit{Stage 2 - Masked video reconstruction}

\vspace{0.2cm}

As Stage 1 performs translation independently at the frame level, temporal consistency across consecutive frames is not explicitly enforced, which may affect the continuity of cardiac motion and image appearance. To address this limitation, Stage 2 introduces a video-based refinement step based on LeanVAE, an ultra-efficient video reconstruction method \cite{b16}. Its lightweight architecture combines non-overlapping patch operations, wavelet transforms, and channel compression to preserve global structure and fine-scale details. In this work, LeanVAE was adapted to masked reconstruction and trained on preprocessed MIMIC-IV-ECHO sequences to learn view-specific spatiotemporal patterns of real cardiac motion.

Consistent with Stage 1, a separate model was trained for each echocardiographic view. During training, fan-shaped regions along the ultrasound beam direction were masked (Fig.~\ref{fig:framework_overview}), encouraging reconstruction of missing regions from the surrounding spatiotemporal context. At inference, one or two fan-shaped masks were positioned over any visible valve region in the Stage 1 outputs, and the full reconstructed video was retained. This encourages the recovery of fine dynamic structures and motion patterns, such as valve dynamics, that may be inconsistently represented or missing, reducing frame-wise translation artifacts and improving temporal coherence.

\section{Experiments and results}
Training was performed on the Deucalion high-performance computing infrastructure using an NVIDIA A100 80 GB GPU.

\subsection{Dataset}
To implement and evaluate the proposed framework, 1,000 dynamic whole-heart meshes were generated using the CONRAD statistical shape model (SSM) \cite{b17}, following \cite{b18}. Anatomical variability was introduced by sampling the statistical modes within $\boldsymbol{\alpha} \in [-3,3]^{16}$ and applying a chi-squared constraint ($\sum_{i=1}^{16}\alpha_i^2\leq\chi^2_{16,0.95}$; otherwise, resample $\boldsymbol{\alpha}$) to exclude implausible anatomical variations. As the SSM explicitly models myocardial tissue only for the left ventricle, wall models were generated for the remaining chambers and the aorta by dilating the corresponding structures. The resulting population spanned an ejection fraction range of 30\%--80\%, providing anatomical and functional variability.

From these meshes, 42,000 echocardiographic sequences were simulated across the seven target views, with each sequence spanning one complete cardiac cycle. For each view, 6,000 sequences were generated, corresponding to six acquisitions per mesh under different simulated conditions.

For the real domain, the preprocessed MIMIC-IV-ECHO dataset comprised 32,663 sequences, distributed across apical (62\%), parasternal (31\%), and subcostal (7\%) views.

The source and target datasets used in both stages were split into training, validation, and test sets using an 80/10/10 partition at the (virtual) patient level. The validation set was used for hyperparameter tuning and model selection, while the held-out test set was reserved for final evaluation.

\subsection{Implementation details}
For Stage 1, the network configuration largely followed \cite{b13} in terms of input size, loss weighting, and optimization settings. The models were trained for 130 epochs using 10 diffusion timesteps, a batch size of 4, and discriminator and generator learning rates of $1\times10^{-5}$ and $1.6\times10^{-5}$, respectively. Data augmentation was applied on-the-fly using random 10\% cropping and horizontal flipping. 

For Stage 2, the LeanVAE architecture followed \cite{b16} and was trained on $256\times256$ sequences of 17 frames uniformly resampled from complete cardiac cycles, using a batch size of 12. The learning rate underwent a 5,000-step warm-up to $5\times10^{-5}$, followed by a gradual decay to $1\times10^{-5}$. Training was fixed at 88,000 steps to account for differences in dataset size across views. The reconstruction objective combined $L_1$ losses over masked and visible regions, weighted by 0.85 and 0.15, respectively, together with a temporal consistency term weighted by 0.5 within the masked regions. During training, the ultrasound sector was identified from its lower corners, and single- or dual-beam masks were randomly positioned within it, each covering 15–40\% of the sector's angular width. At inference, dual-beam masks were used when two target valve regions could be covered without overlap; otherwise, a single-beam mask was applied. Beam widths were adapted to encompass the target valve regions while remaining within the training range. For shortened views in which no valve could be reliably localized, randomly positioned masks were used.

For both stages, model selection was based on a combined assessment of image realism and anatomical preservation. Realism was assessed using the Fréchet Inception Distance (FID) \cite{b19}, computed from pretrained EchoPrime features \cite{b15} extracted from generated images and corresponding MIMIC-IV-ECHO validation data. Anatomical preservation was assessed using the Spatially-Correlative Loss (SCL) \cite{b20}, which evaluates structural consistency by comparing spatial relationships within generated and source images. For SCL computation, only anatomical regions were considered, and Gaussian blur ($\sigma=3$) was applied to reduce sensitivity to sharp transitions introduced by the label-to-intensity mapping. The final selection score was defined as $\mathrm{FID}/1000+\mathrm{SCL}$, scaling FID to a range comparable to SCL. The checkpoint with the lowest combined score was selected.

\subsection{Impact of the two-stage approach}
\begin{figure*}[!t]
    \centering
    \includegraphics[width=0.85\textwidth]{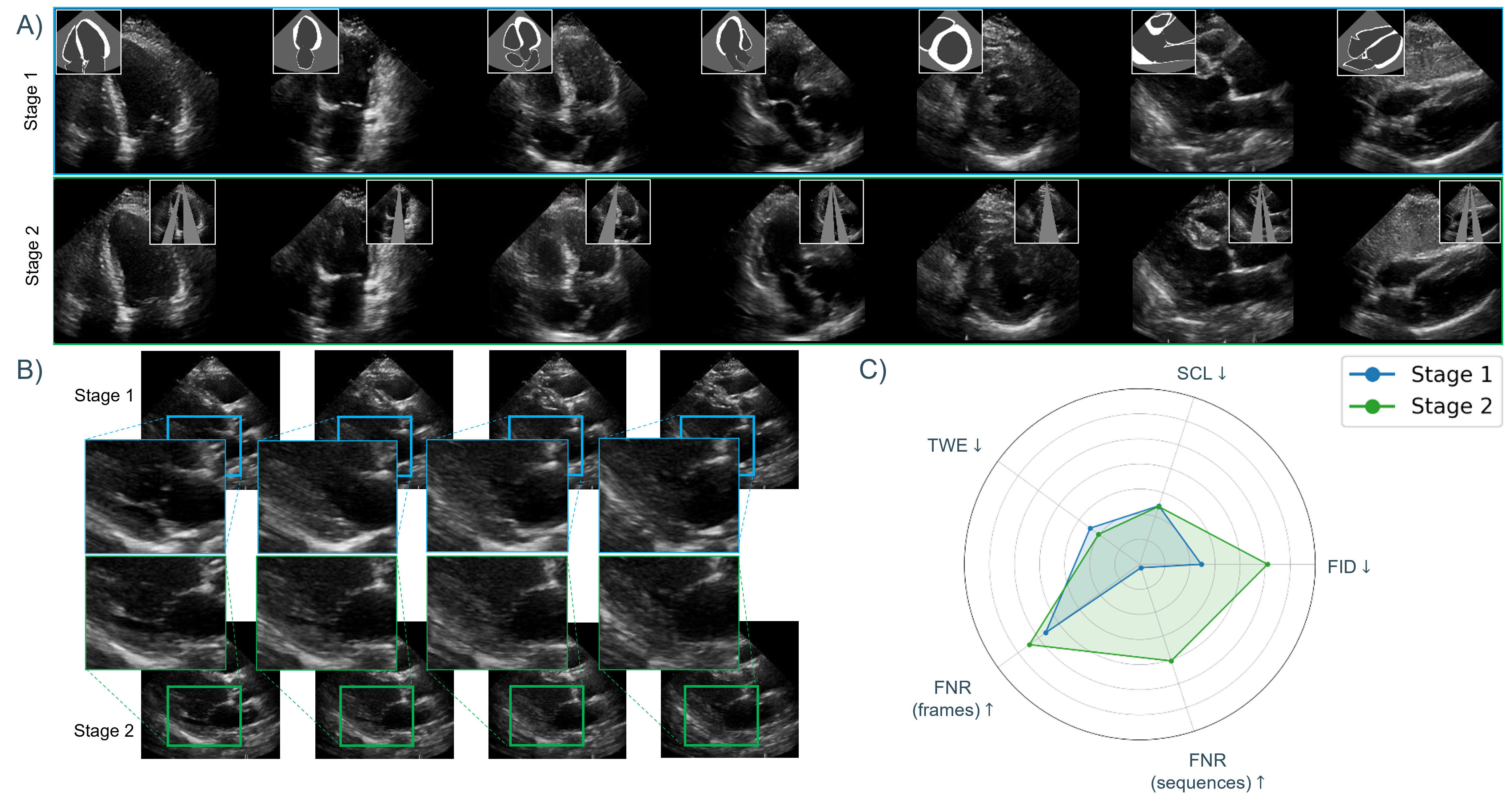}
    \caption{Qualitative and quantitative evaluation of the proposed framework. (A) Representative examples across different echocardiographic views, showing Stage~1 and Stage~2 outputs alongside the corresponding simulated anatomy. (B) Four consecutive frames with magnified regions highlighting improved valve and wall dynamics and reduced temporal artifacts after Stage~2. (C) View-averaged FID, SCL, TWE, and blinded reader-study results on the test set.}
    \label{fig:results}
\end{figure*}

The two-stage framework demonstrated complementary improvements in image realism and temporal consistency. Stage~1 introduced realistic ultrasound appearance across all views (Fig.~\ref{fig:results}A), while Stage~2 refined valve and wall dynamics, reducing local artifacts and frame-to-frame jitter to promote smoother temporal evolution (Fig.~\ref{fig:results}B).

To evaluate these improvements, image realism, anatomical preservation, and temporal consistency were assessed using EchoPrime-based FID, SCL, and temporal warping error (TWE) \cite{b21}, respectively. These metrics were complemented by a blinded reader study in which two cardiologists (A.Y. and V.H.P.) classified end-diastolic frames and full sequences as real or synthetic to assess frame- and video-level perceptual realism. Each evaluation set comprised 210 samples: 70 real MIMIC-IV-ECHO examples and 70 synthetic examples from each stage (10 per view). Synthetic samples were selected from near-optimal transducer poses to minimize discrimination based on atypical anatomy. Reader responses were summarized using false negative rate (FNR; proportion of synthetic samples classified as real). Results are presented in Fig.~\ref{fig:results}C.

Stage~1 achieved lower FID, indicating greater similarity to real echocardiographic data in the feature space, whereas Stage~2 reduced TWE, indicating improved temporal consistency. Similar SCL values suggested comparable anatomical preservation across both stages. Despite its higher FID, Stage~2 achieved greater perceptual realism in the blinded reader study, with reader-averaged FNR increasing from 66.43\% to 77.86\% for frames and, more markedly, from 2.14\% to 57.86\% for sequences. These findings highlight the contribution of video-level modeling to perceptual realism through improved valve and wall dynamics. Nevertheless, FNR decreased from frames to sequences in both stages, accompanied by a reduction in the false positive rate (proportion of real samples classified as synthetic) from 20.71\% to 4.29\%, suggesting that temporal information facilitated discrimination between real and synthetic data. This may partly reflect limitations of the SSM, including the limited temporal resolution of the original meshes and constrained cardiac dynamics, potentially producing motion patterns that differ from clinical echocardiography. Evaluating the framework using single-patient meshes with higher temporal resolution could help assess its performance with more representative cardiac motion.

\section{Conclusion}
We presented a novel framework combining anatomically consistent multi-view simulation with two-stage synthesis to generate realistic, shape-controlled echocardiographic sequences. Quantitative and expert evaluations demonstrated complementary improvements in image realism and temporal consistency, highlighting the benefits of both synthesis stages. By enabling anatomically controlled generation across multiple views, the framework may complement clinical datasets and facilitate the development of comprehensive patient-level cardiac assessment methods.

\end{document}